\documentclass[conference]{IEEEtran}
\IEEEoverridecommandlockouts
\usepackage{amsmath,amssymb,amsfonts}
\usepackage{algorithm,algorithmic}
\usepackage{graphicx}
\usepackage{orcidlink}
\usepackage[caption=false,font=footnotesize]{subfig}
\usepackage{tabularx}
\usepackage{array}
\usepackage[table,dvipsnames]{xcolor}   
\usepackage{multirow}
\usepackage{cite}                        
\usepackage{hyperref}                    
\hypersetup{
    colorlinks=true,
    linkcolor=blue,
    citecolor=blue,
    urlcolor=blue,
    pdfborder={0 0 0},
    breaklinks=true
}



\newcolumntype{Y}{>{\centering\arraybackslash}X}

\begin{document}
\title{PALTO: Physics-Informed Active Learning for Tri-Gate FinFET Design Optimization for Vertical Power Delivery}

\author{%
  \IEEEauthorblockN{Ayoub Sadeghi\,\orcidlink{0000-0001-9904-9813}\\
\textit{University of Illinois Chicago}\\
Chicago, USA\\
asadeg3@uic.edu
\vspace{-20pt}}
  \and
  
 \IEEEauthorblockN{Leonid~Popryho\,\orcidlink{0009-0002-0578-9592}\\
\textit{University of Illinois Chicago}\\
Chicago, USA\\
lpopry2@uic.edu
\vspace{-20pt}} 
  \and
  \IEEEauthorblockN{Inna~Partin\mbox{-}Vaisband\,\orcidlink{0000-0002-6399-6672}\\
\textit{University of Illinois Chicago}\\
Chicago, USA\\
vaisband@uic.edu
\vspace{-20pt}} 
    \thanks{%
    
    This work was supported by the Center for Heterogeneous Integration of Micro Electronic Systems (CHIMES), one of seven centers in Joint University Microelectronics Program (JUMP) 2.0, a Semiconductor Research Corporation (SRC) program sponsored by the Defense Advance Research Project Agency (DARPA). We thank Prof. Debjit Pal for providing the distributed simulation infrastructure.
  }%
}
\maketitle
%
\begin{abstract}
This paper demonstrates the effectiveness of machine learning-driven optimization for designing application-specific GaN tri-gate FinFETs in vertical power delivery systems. Conventional TCAD-based approaches are computationally intensive and insufficient for navigating the high-dimensional, nonlinear design space of advanced GaN devices. To address this, a physics-informed active learning framework is used to intelligently guide simulations, accelerating convergence while preserving accuracy. This ML-guided approach enables the discovery of optimal configurations by efficiently exploring key structural parameters---most notably the GaN-to-AlGaN thickness ratio---a long-standing focus of debate in device design.
By systematically exploring key structural parameters,
two optimized devices with aggressively scaled gate-to-drain lengths are identified. Single-fin, multi-channel simulations show that device~D2, with a thinner GaN channel relative to the AlGaN barrier, achieves higher drive current. However, in a 300-fin configuration, device~D1 outperforms device~D2 by delivering 3.3\,A at 0.49~ohm on-resistance---approximately 2$\times$ better---despite slightly higher parasitics. Both devices operate in a normally-off mode. Based on an application-specific figure of merit, device~D1 achieves 5\,pC$\cdot$ohm, demonstrating 2$\times$ greater switching efficiency than device~D2, while both designs outperform industrial benchmarks from different performance standpoints.
\end{abstract}


\vspace{-5pt}
\section{Introduction}
\label{sec:introduction}

\begin{figure*}[t]
    \centering  
    \vspace{-10pt}
    \includegraphics[width=\linewidth]{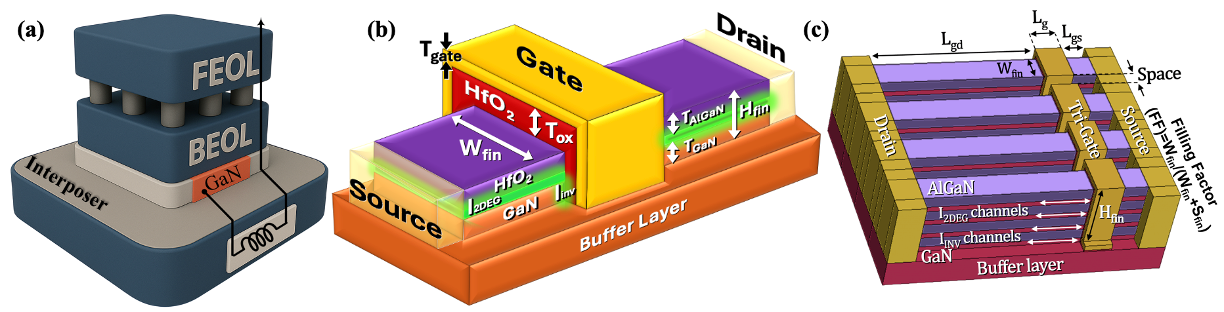}    
    \caption{Vertical power delivery, (a) system architecture with GaN power switches integrated beneath the functional die~\cite{ECTC}, (b) tri-gate FinFET with a single fin and single channel, illustrating key design parameters, and (c) multi-fin configuration of the tri-gate FinFET with four parallel channels.}
    \label{fig:Finfet-App}
\end{figure*}

The growing demand for efficient power delivery in next-generation computing platforms has driven the adoption of vertical power delivery (VPD) architectures, where power switches are strategically integrated directly beneath the functional die to minimize interconnect losses and improve power density (see Fig.~\ref{fig:Finfet-App}(a))~\cite{Sriharini,krishnakumar2023vertical}. To effectively enable VPD approaches, compact, high-performance power switches should be embedded in close proximity to points-of-load (POLs), thereby reducing energy losses and enhancing system efficiency~\cite{ECTC, EPEPS-salma, Rami-ECTC, Dula-GLSVLSI, Dula-ECTC, glsv}.

Among the various device candidates, multi-fin, multi-channel tri-gate GaN FinFETs have emerged as a highly promising solution due to their superior scalability and ability to sustain high current densities~\cite{main-multifin, ECTC, optimization,Fin-shape,nature-revi,nela2-FF,single-mA,pgate}. Their performance builds upon the capabilities demonstrated by single-fin, multi-channel GaN FinFET units, which serve as the foundational elements of the design (see Fig.~\ref{fig:Finfet-App}(b)(c))~\cite{main-multifin, ECTC}.

Albeit the initial promise, GaN switches that meet the stringent VPD performance requirements in modern systems-on-package (SoPs) and chiplet-based systems have not yet demonstrated. Particularly, performance targets include  
large voltage conversion ratios (e.g., stepping down from 48~V or 12~V to 1~V), current densities exceeding 2--4~A/mm\textsuperscript{2}, enhancement mode (E-mode) operation, and conversion efficiencies in the range of 90\%--93\%. To achieve the required efficiency, switching and conduction losses should be carefully balanced, posing additional design constraints on parasitic resistance and capacitances in such devices~\cite{ECTC-salma}.

Optimization of complex multi-fin multi-channel devices relies on precise co-design of key structural single-channel device parameters, requiring numerous iterations of extensive simulations~\cite{ECTC, optimization, I2deg-Iinv, single-mA} (see Fig.~\ref{fig:Finfet-App}(b)). Traditional TCAD-based exploration is, however, computationally expensive and poorly suited for navigating the highly nonlinear, high-dimensional design space~\cite{ECTC, optimization}. These limitations often result in suboptimal solutions, especially for non-planar GaN devices in VPD systems. Instead, intelligent, data-efficient optimization strategies and design space exploration techniques are required to efficiently navigate complex, high-dimensional parameter landscapes and identify optimal device configurations.

\begin{figure*}[t]
    \centering        \includegraphics[width=\linewidth]{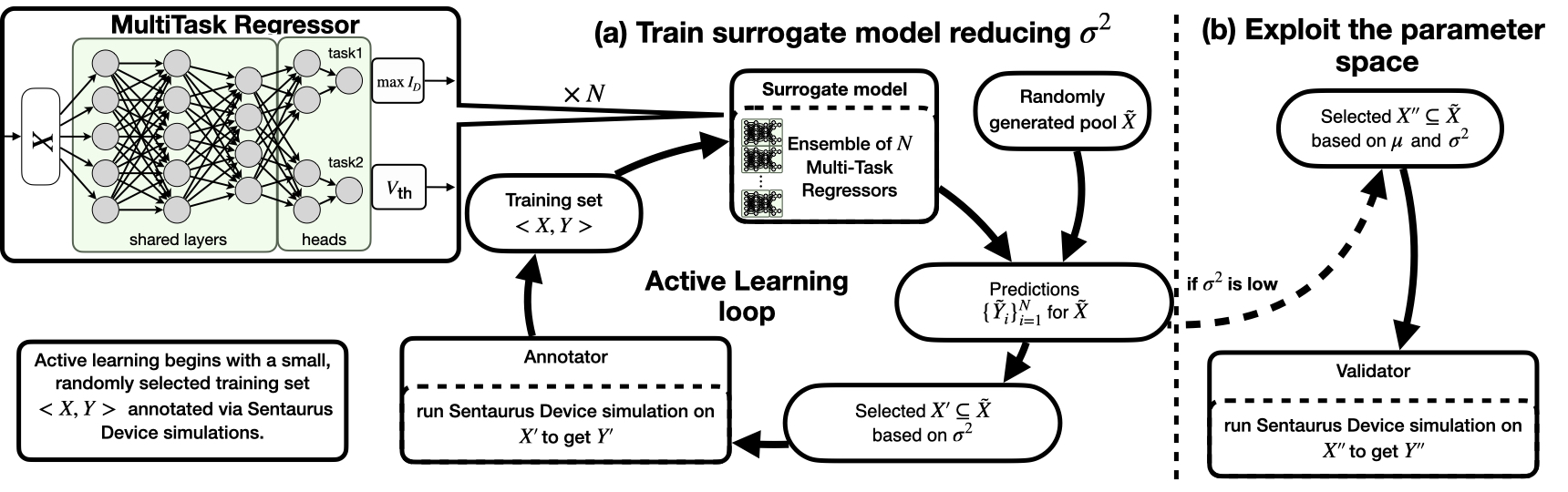}    
    \caption{PALTO active-learning workflow, (a) exploration (i.e., candidate selection and labeling using an ensemble of multi-task regressors), and (b) exploitation (i.e., selection of high-performance, low-uncertainty designs using the trained model).}
    \vspace{-15pt}
    \label{fig:Palto-framework}
\end{figure*}

To address this challenge, a physics-informed machine learning (ML) framework is proposed in this paper. The workflow of the framework is illustrated in {Fig.~\ref{fig:Palto-framework}}. The framework was developed and trained on a systematically generated TCAD dataset. The model effectively captures complex, nonlinear dependencies between structural device parameters and performance metrics, enabling rapid, targeted exploration of the design space. Compared to conventional methods, the ML-guided approach significantly reduces simulation time and computational overhead, while providing valuable physical insights into how critical features---such as layer thicknesses, channel geometry, and doping profiles---impact switching efficiency~\cite{nela-nature}. Additional insight is gained by calculating SHAP (SHapley Additive exPlanations) values, a game theory–based method that quantifies how much each input feature (e.g., a device parameter or design variable) contributes to the model’s output or decision (see~Fig.~\ref{fig:shap-values}).

By targeting high output current ($I_{\text{DS}}$) and a positive threshold voltage ($V_{\text{th}}$), the ML-guided optimization process identifies two distinct device configurations, named as device~D1 and device~D2, primarily differentiated by the thicknesses of the GaN channel and AlGaN barrier---a primary challenge among similar device optimizations~\cite{opt-device}. Device~D1 features a structure with thicker GaN, while device~D2 employs a thicker AlGaN barrier relative to the GaN channel. When scaled to a 300-fin architecture (see {Fig.~\ref{fig:Finfet-App}(c)}), device~D1 achieves $R_{\text{on}} = 0.49~\Omega$ and $I_{\text{DS}} = 3.3~\text{A}$, compared to $0.94~\Omega$ and 1.67~A for device~D2---both outperforming the non-ML-optimized baseline~\cite{ECTC} and commercial benchmarks. Additionally, both designs exhibit low parasitic capacitances, offering a scalable and high-performance solution for integration of GaN power transistors into next-generation compute platforms.

\section{PALTO Methodology}
\begin{figure}[t]
    \centering
    \vspace{-5pt}
    \includegraphics[width=0.9\linewidth]{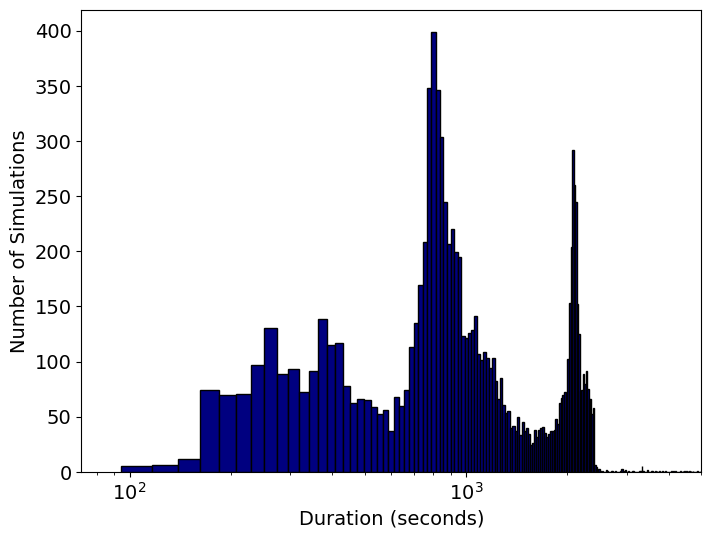}
    \vspace{-5pt}
    \caption{Distribution of TCAD simulation runtimes across all completed trials.  
    Individual runs span from 1.5\,min to 6\,h\,15\,min, with a heavy-tailed spread and a median of $\sim$16\,min.}
    \label{fig:durations}
\end{figure}

\vspace{-10pt}
While TCAD simulations offer accurate device-level physics, a TCAD-based evaluation of a set of device parameters requires on average 16 minutes and up to 6.25 hours for those more complex nonlinear settings (see {Fig.~\ref{fig:durations}}). Brute-force exploration of such design space is, therefore, impractical. To overcome this, the \textbf{PALTO} (Physics-Informed Active Learning for Tri-Gate FinFET Optimization) framework is introduced. It couples a query-by-committee active learning loop with a deep-ensemble surrogate model, prioritizing high-value candidates for TCAD evaluation while screening millions of low-uncertainty variants in seconds.

Each simulation is an expensive to evaluate function
$f_{\text{TCAD}}:\mathbf{x}\mapsto\{I_{\text{DS}\mathrm{max}},V_{\text{th}}\},$
where $\mathbf{x}$ covers the ranges in Table~\ref{tab:Optimizarion-Ranges}.  
An initial Latin Hypercube Sampling (LHS)~\cite{latin-hypercube}—a statistical method for generating representative samples from a multidimensional parameter space—of about $10^{3}$ different device configurations (guided by~\cite{ECTC,nela-nature}) with corresponding simulated $I_{\text{d,max}}$ and $V_{\text{th}}$ results is used to bootstrap the surrogate model.

\begin{figure}[t]
    \centering
    \hspace{-15pt}
    \includegraphics[width=\linewidth]{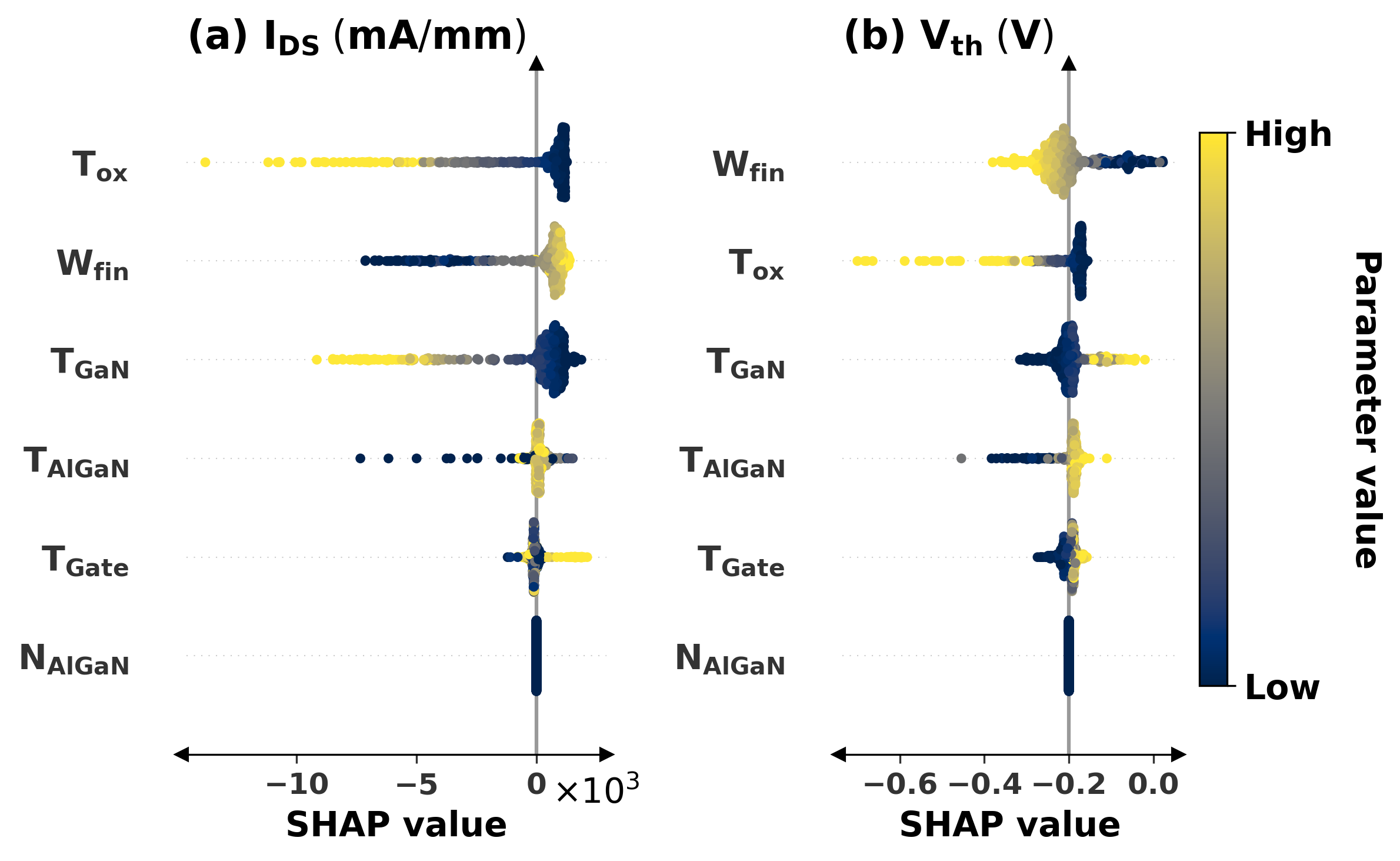}
    \caption{SHAP beeswarm plots for (a) $I_{\text{DS,max}}$ and (b) $V_{\text{th}}$, where positive (negative) values indicate increasing (decreasing) contribution of device parameters. The plots reveal that $T_{\text{ox}}$ and $W_{\text{fin}}$ dominate both metrics, with opposing trends between them.}
    \label{fig:shap-values}
\end{figure}

A multitask neural network with a shared fully-connected backbone and two task-specific heads simultaneously predicts $I_{\text{DS}\mathrm{max}}$ and $V_{\text{th}}$.  
To balance the multi-objective loss function, training uses homoscedastic-uncertainty weighting, a technique that balances multiple loss terms by modeling each task’s constant output uncertainty~\cite{homoscedastic-uncertainty}.

{\small
\begin{equation}
    \mathcal{L}=\frac{\|I_{\mathrm{DS,max}}-\hat{I}_{\mathrm{DS,max}}\|^{2}}{2\sigma_{I}^{2}}+
  \frac{\|V_{\mathrm{th}}-\hat{V}_{\mathrm{th}}\|_{w}^{2}}{2\sigma_{V}^{2}}+
  \log\sigma_{I}+\log\sigma_{V}
\end{equation}
}
with an additional ten-fold penalty applied to errors where $V_{\text{th}}>0$ in order to maintain high accuracy in the enhancement-mode regime.  
An ensemble of $N=10$ such networks captures epistemic uncertainty, which reflects the model’s lack of knowledge due to limited training data, finite model capacity, and imperfect underlying assumptions,

\begin{equation}
  \mathcal{U}(\mathbf{x})=\operatorname{Var}\!\bigl\{M_{i}(\mathbf{x})\bigr\}_{i=1}^{N}.
\end{equation}
\vspace{-15pt}

After convergence ($R^{2}>0.98$ on both tasks), the ensemble is able to predict $6\times10^{9}$ designs in about~10~minutes on a single RTX~4090, while only around $10^{4}$ TCAD calls are required for convergence in total.  
All runs, metrics, and artifacts are tracked with Weights~\&~Biases (W\&B)—a platform for experiment tracking, hyperparameter tuning, and reproducibility—to ensure that the modeling workflow is reproducible, well-logged, and version-controlled, thus improving transparency and auditability of results.

The algorithm adopts a query-by-committee strategy (see {Algorithm~\ref{alg:palto}}), where deep-ensemble is used to estimate epistemic uncertainty and guide the selection of informative samples~\cite{al-literature-review,deep-ensembles}.  

\begin{algorithm}[ht]
  \caption{PALTO active-learning loop}\label{alg:palto}
  \begin{algorithmic}[1]
    \STATE The labeled pool $\mathcal{D}_{0}$ is initialized with $k$ uniformly distributed simulations
    \FOR{round $t=1$ until validation uncertainty stabilizes}
      \STATE The deep-ensemble surrogate $\{M_{i}\}_{i=1}^{N}$ is trained on $\mathcal{D}_{t-1}$
      \STATE A large candidate pool of size $n$ is drawn over the design space and the uncertainty score $\mathcal{U}(\mathbf{x})$ is evaluated
      \STATE Candidates are ranked by $\mathcal{U}$; a greedy farthest-first filter selects a diverse subset of $m$ points
      \STATE High-fidelity TCAD simulations are executed for these $m$ and the results are appended to form $\mathcal{D}_{t}$
    \ENDFOR
  \end{algorithmic}
\end{algorithm}

\textbf{\textit{Experimental setup.}}
In the reported study, the initial pool comprised approximately $k=10^{3}$ uniformly sampled devices; each iteration assessed $n=10^{6}$ randomly drawn candidates, from which batches of $m=100$ designs were simulated with Sentaurus before retraining. The loop terminated once the ensemble variance on a held-out validation grid stops decreasing.

Prediction variance alone tends to cluster similar candidates.  
Before each TCAD batch, a greedy farthest-first pass is applied to maximize the minimum pair-wise $L_{2}$ distance in the parameter space, pruning redundant candidates without information loss.

SHAP beeswarm plots ({Fig.~\ref{fig:shap-values}}) computed on the converged ensemble with \texttt{shap.DeepExplainer} are utilized to rank the dominant features for \textbf{(a)} $I_{\text{DS}\mathrm{max}}$ and \textbf{(b)} $V_{\text{th}}$.  
Each point represents an deck (i.e., a set of input parameters); color encodes the raw feature value, and the x-axis shows the signed contribution relative to a reference.

\begin{figure}[t]
    \centering
    \vspace{-10pt}
    \includegraphics[width=\linewidth]{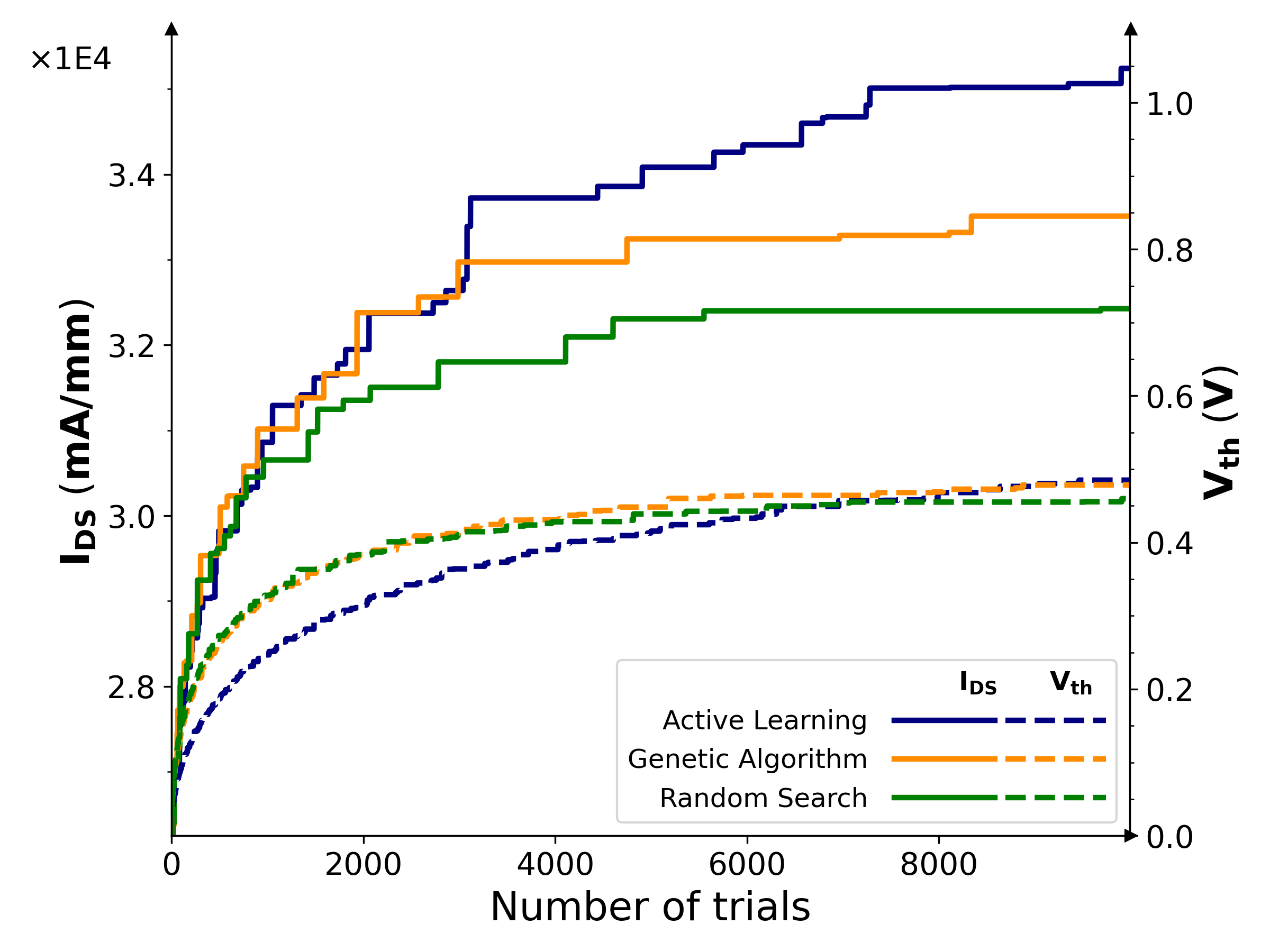}
    
    \caption{Convergence of best $I_{\text{DS}\mathrm{max}}$ (solid) and $V_{\text{th}}$ (dashed) versus cumulative TCAD calls; PALTO compared with NSGA-II and random search.}
    \label{fig:learning-curve}
    \vspace{-5pt}
\end{figure}

\begin{figure}[t]
    \centering
    \hspace{-20pt}
    \includegraphics[width=\linewidth]{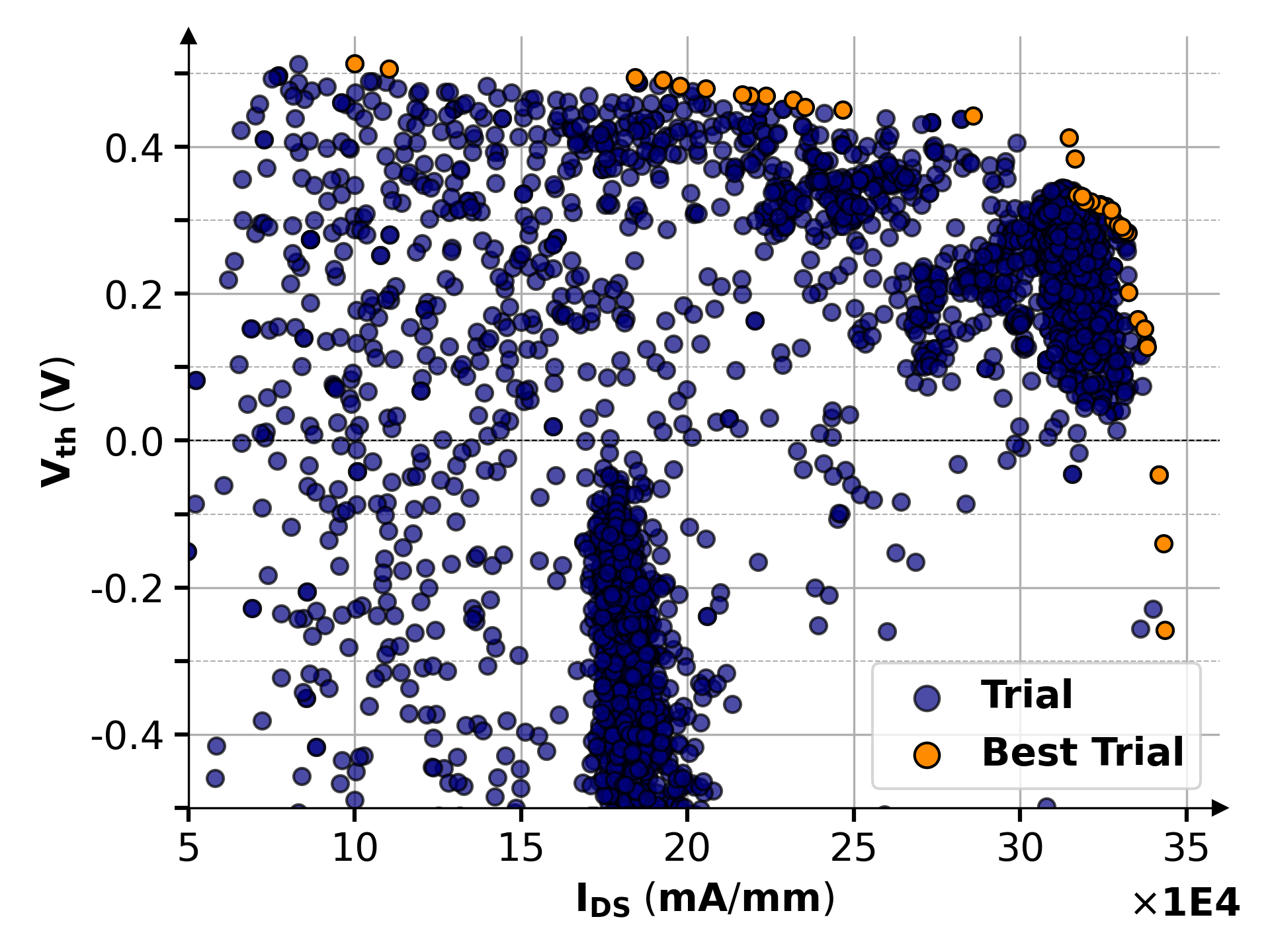}
    \vspace{5pt}
    
    \caption{Simulated performance for all explored tri-gates. Pareto front clusters along the upper-right boundary, indicating simultaneous improvement$I_{\text{DS}\mathrm{max}}$ and $V_{\text{th}}$ metrics within the explored design space.}
    \label{fig:pareto-front}
\end{figure}

Compared with NSGA-II~\cite{nsga2}, a widely used multi-objective evolutionary algorithm for identifying Pareto-optimal solutions under conflicting objectives, used here to maximize $I_{\text{DS}}$ under the constraint $V_{\text{th}} > 0$, and random search, PALTO achieves a $3.2\times$ reduction in TCAD cost to reach equivalent or superior device performance (see Fig.~\ref{fig:learning-curve}).
Based on the learned design landscape (Fig.~\ref{fig:pareto-front}), two devices meeting the design targets are identified for comprehensive DC and AC evaluation, in 50-device multi-fin parallel configuration.
 
\section{Device Structure and Results}

A state-of-the-art baseline configuration~\cite{ECTC} and the two identified optimal devices (see {Table~\ref{tab:Optimizarion-Ranges}} for a list of key design parameters) share a gate length ($L_{\text{g}}$) of 0.1~\textmu m and a gate-to-drain spacing ($L_{\text{gd}}$) of 0.6~\textmu m, resulting in a total device length of approximately 1~\textmu m. Reducing $L_{\text{gd}}$ enhances the drive current and lowers the on-resistance ($R_{\text{on}})$, albeit at the cost of reduced breakdown voltage (BV). However, for the targeted power delivery application, a BV of above 50~V remains acceptable. All devices employ HfO\textsubscript{2} as the gate dielectric and utilize an AlGaN barrier layer with a composition of Al\textsubscript{0.25}Ga\textsubscript{0.75}N. Device parameters are normalized to the effective width ($W_{\text{eff}}$~=~$2H_{\text{fin}}$~+~$W_{\text{fin}}$) for consistent performance comparison. The optimization intervals and the final parameter values of the individual devices are listed in {Table~\ref{tab:Optimizarion-Ranges}}, highlighting the structural differences between the two optimized configurations. All simulations are performed using 3D TCAD Sentaurus. To ensure accurate representation of electrostatics and carrier transport, advanced physical models, such as Shockley-Read-Hall (SRH) recombination, interface trap modeling, and finite element methods (FEM) dynamics are incorporated~\cite{optimization, ECTC}.

\begin{table}[htbp]
    \centering
    \caption{\centering Key Design Parameters and Their Respective Intervals Used in the Exhaustive Simulation Procedure, Compared Against Those Selected by the ML-Based Method for Optimal Device Identification.}
    \renewcommand{\arraystretch}{1.3}
    \setlength{\tabcolsep}{4.2pt}

    \begin{tabularx}{\linewidth}{|l|c|c|c|c|}
        \hline
        \multirow{3}{*}{\textbf{Description}} & 
        \textbf{Extensive} &\multicolumn{3}{c|}{\textbf{ML-Based}} \\ \cline{3-5}
         & \textbf{Simulation} & \textbf{Optimization} & \multicolumn{2}{c|}{\textbf{Selected Devices }} \\ \cline{4-5}
         & \textbf{Baseline~\cite{ECTC}} & \textbf{Interval} & \textbf{\text{D1}} & \textbf{\text{D2}} \\ \hline \rowcolor{gray!20}
        Doping ($10^{18}\text{cm}^{-3}$) 
        & $1$ & $1~...~100$ & $17.1$ & $94$ 
        \\ \hline
        Fin width (nm) 
        & $100$ & $10~...~250$ & $87$ & $81$ \\ \hline \rowcolor{gray!20}
        GaN thickness (nm) 
        & $20$ & $10~...~100$ & $99$ & $11$ \\ \hline
        AlGaN thickness (nm) 
        & $40$ & $10~...~100$ & $83$ & $95$ \\ \hline \rowcolor{gray!20}
        Oxide thickness (nm) 
        & $6$ & $5~...~25$ & $6$ & $5$  \\ \hline
        Gate thickness (nm) 
        & $20$ & $5~...~40$ & $5$ & $37$ \\ \hline \rowcolor{gray!20}
        Number of channels 
        & $4$ & $4$ & $4$ & $4$ \\ \hline
    \end{tabularx}
    \label{tab:Optimizarion-Ranges}
\end{table}

\begin{figure*}[t]
    \centering  
    \vspace{-10pt}
    \includegraphics[width=\linewidth]{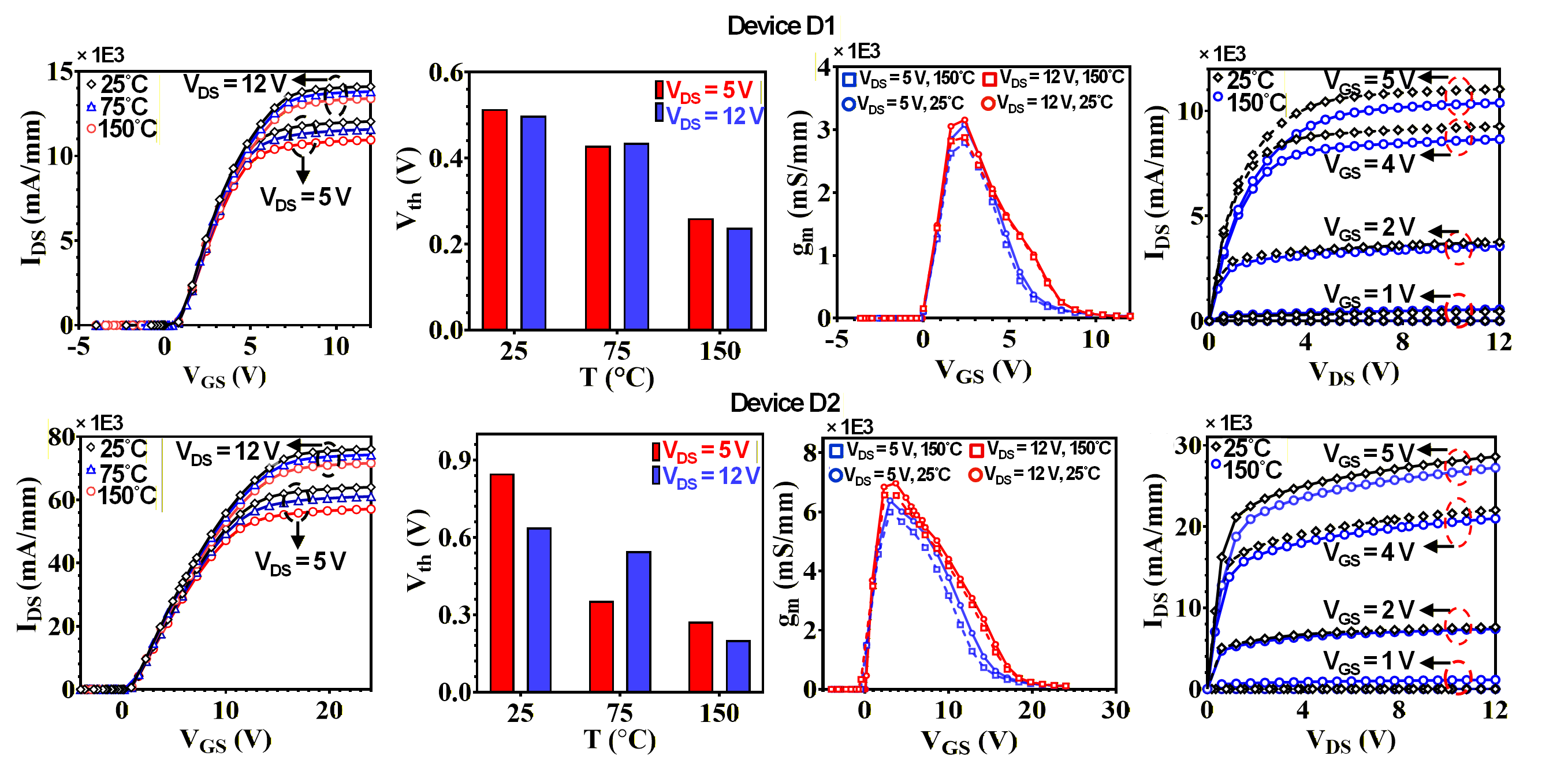}    
    \caption{Temperature-dependent performance of the optimized single-fin, multi-channel devices~D1 and~D2 with $L_{\text{gd}} = 0.6~$\textmu m and $L_{\text{g}} = 0.1~$\textmu m. The results demonstrate stable operation across the voltage and temperature ranges of interest.}
    \label{fig:Output-Chract}
\end{figure*}

\textbf{\textit{Single-Fin Configuration.}}
Temperature-dependent transfer characteristics ($I_{\text{DS}}$-$V_{\text{GS}}$) of the two optimized single-fin four-channel devices are shown in {Fig.~\ref{fig:Output-Chract}} for variety of $V_{\text{DS}}$ values. As temperature increases, $I_{\text{DS}}$ decreases and $V_{\text{th}}$ shifts negatively, as further illustrated in {Fig.~\ref{fig:Output-Chract}}. Despite this negative shift, both devices maintain E-mode operation, demonstrating strong thermal stability.

Specifically, device~D1 exhibits a peak current of 14,101~mA/mm at 25\textdegree{C} and ${V_{\text{DS}} = 12}~V$, which is approximately 14\% higher than its performance at ${V_{\text{DS}} = 5}$~V. It also shows fast current saturation and minimal degradation with increasing temperature. Device~D2 achieves a significantly higher peak current of 76{,}105~mA/mm at 25\textdegree{C} and ${V_{\text{DS}} = 12}$~V, featuring a wider $V_{\text{GS}}$ range that makes it well-suited for linear-region operation---a preferred mode for power switching---while maintaining similarly low temperature-induced degradation.
Device~D1 exhibits a total $V_{\text{th}}$ shift of 0.25~V from 25\textdegree{C} to 150\textdegree{C} at ${V_{\text{DS}} = 5}$~V and 12~V, indicating good thermal stability. In contrast, device~D2 shows a slightly larger shift of 0.57~V at ${V_{\text{DS}} = 5}$~V and 0.43~V at ${V_{\text{DS}} = 12}$~V, attributed to its higher gate-to-oxide thickness ratio. Yet $V_{\text{th}}$ remains positive across temperature and bias voltages, confirming the normally-off behavior of both devices. 
Overall, device~D2 delivers a higher $I_{\text{DS}}$ and maintains a more stable $V_{\text{th}}$ across a wider $V_{\text{GS}}$ range, making it more suitable for high-temperature applications.

To evaluate the high-frequency suitability of scaled-down $L_{\text{gd}}$ tri-gate devices, Fig.~\ref{fig:Output-Chract} presents the extracted at $V_{\text{DS}} = 5$~V and 12~V transconductance ($g_{\text{m}}$). 
Device~D2 exhibits a higher peak $g_{\text{m}}$, reaching up to 7,000~mS/mm at 25\textdegree{C} and maintaining 6,500~mS/mm at 150\textdegree{C}. This indicates strong gate control and suitability for high-frequency operation. Device~D1 demonstrates a peak $g_{\text{m}}$ of approximately 3,000~mS/mm with minimal temperature degradation, showing stable transconductance across the operating range. These results confirm that device~D2 offers enhanced performance, comparable to similarly sized reported devices~\cite{optimization, ECTC, single-mA}.

The $I_{\text{DS}}$-$V_{\text{DS}}$ output characteristics of device~D1 and device~D2 are shown for $L_{\text{gd}}$~=~ 0.6~\textmu m in {Fig.~\ref{fig:Output-Chract}} as a function of $V_{\text{GS}}$ and temperature. Device~D1 reaches a maximum $I_{\text{DS}}$ of 11,039~mA/mm at ${V_{\text{DS}} = 12}$~V and ${V_{\text{DS}} = 5}$~V, while device~D2 achieves a significantly higher peak of 28,609~mA/mm. Thus, device~D2 consistently delivers higher $I_{\text{DS}}$ across all gate biases and temperature conditions, demonstrating superior current drive and improved thermal robustness. Both devices exhibit early current saturation and reduced temperature sensitivity at higher gate overdrives.

\begin{figure}[t]
    \centering
    \subfloat{\includegraphics[width=.49\columnwidth]{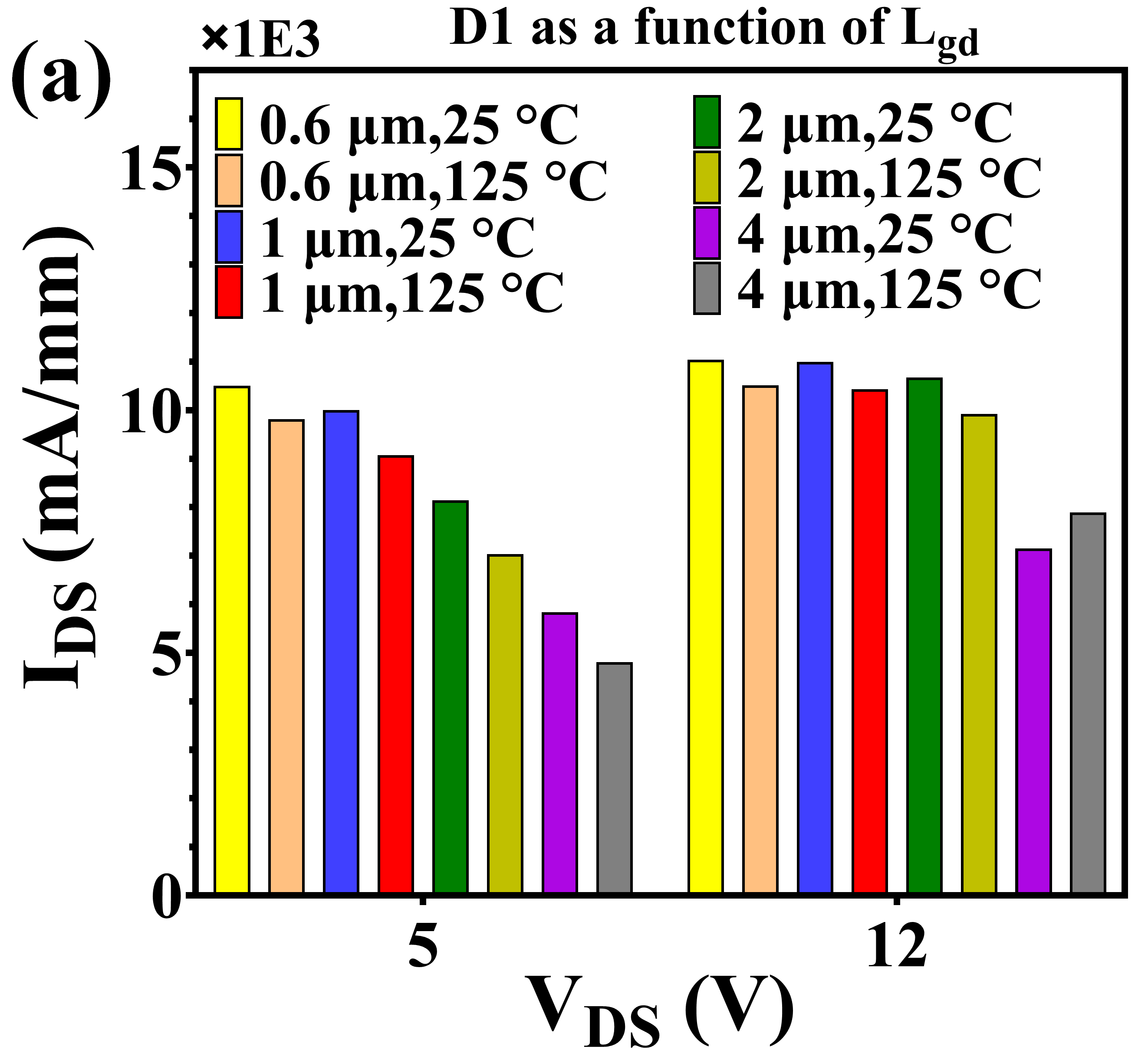}}\label{fig:9a}
    \hfill
    \subfloat{\includegraphics[width=.49\columnwidth]{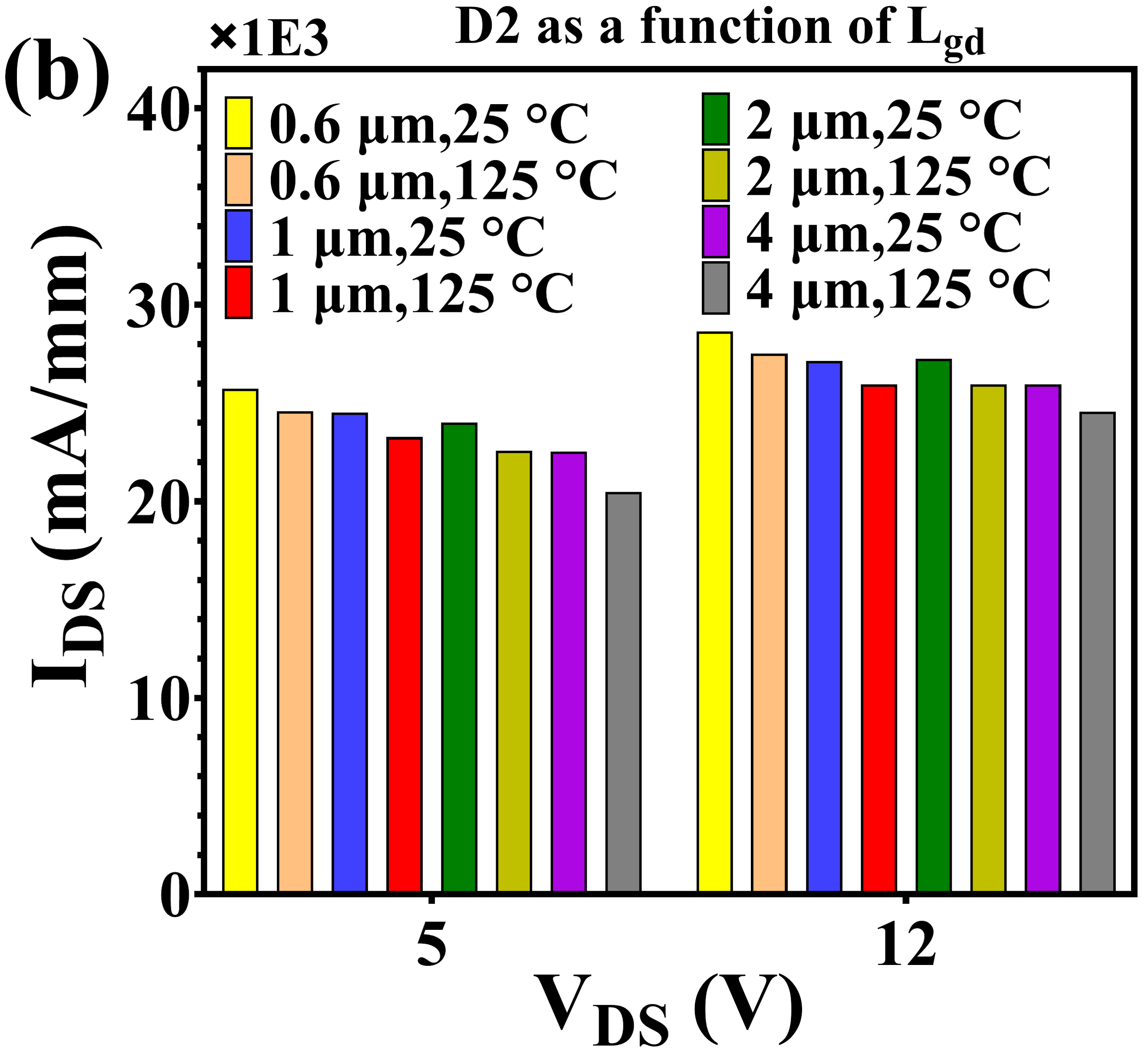}}\label{fig:9b}\\
    \subfloat{\includegraphics[width=.49\columnwidth]{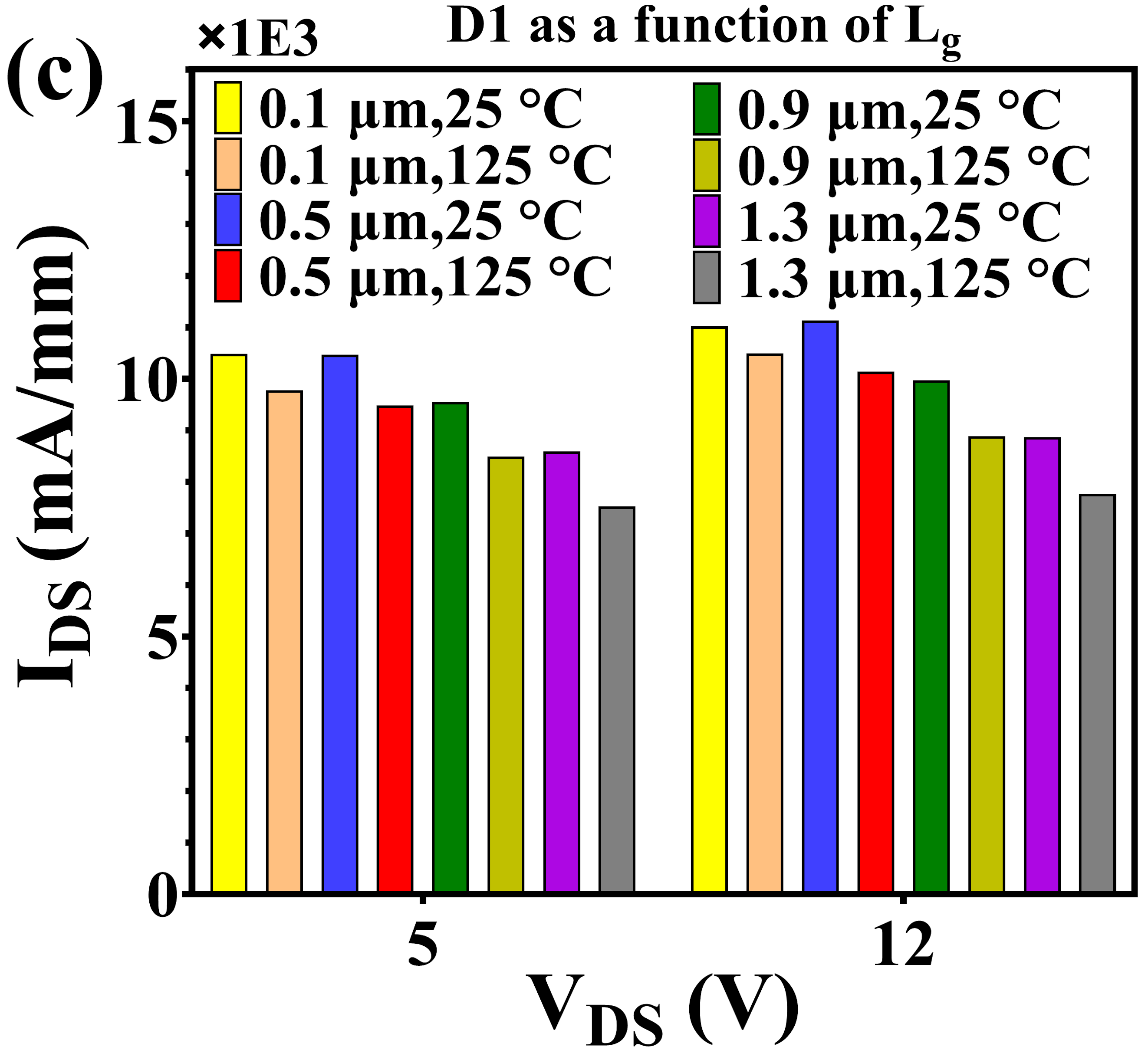}}\label{fig:9c}
    \hfill
    \subfloat{\includegraphics[width=.49\columnwidth]{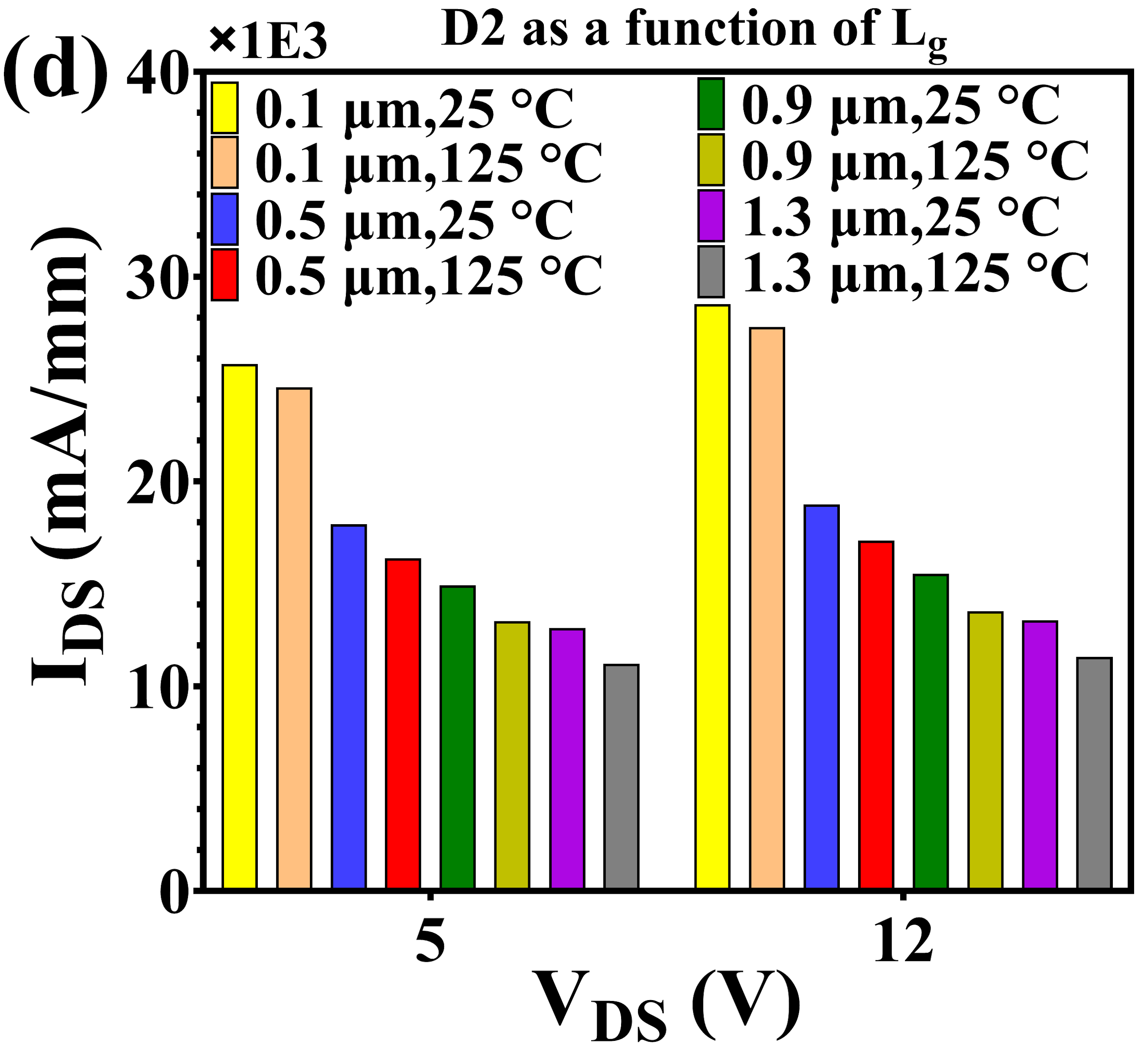}}\label{fig:9d}
\caption{impact of $L_{gd}$ and $L_g$ on maximum current for devices~D1 and~D2, showing that larger $L_{gd}$ and $L_g$ reduce $I_{DS}$, more significantly in D2 and at 125~$^\circ$C.}
    \label{fig:ID-LGD}
    \vspace{-0pt}
\end{figure}

The impact of aggressive gate length ($L_{\text{g}}$) scaling and gate-to-drain ($L_{\text{gd}}$) spacing is shown in {Fig.~\ref{fig:ID-LGD}}, yielding a consistent increase in $I_{\text{DS}}$ with decreasing device dimensions across temperature conditions and $V_{\text{DS}}$ values. Notably, at ${V_{\text{DS}} = 12}$~V, device~D2, with $L_{\text{g}}$ = 0.1~\textmu m and $L_{\text{gd}}$ = 0.6~\textmu m, achieves peak $I_{\text{DS}}$ of 28,609~mA/mm, compared to 24,565~mA/mm at $L_{\text{g}}$ = 0.1~\textmu m and $L_{\text{gd}}$ = 4~\textmu m, maintaining strong performance even at elevated temperatures. The high current drive capability combined with manageable thermal degradation highlights the potential of such aggressively scaled designs for high-performance power applications.

Electrical and small-signal analyses comparing the proposed and baseline~\cite{ECTC, Fin-shape} devices are summarized in {Table~\ref{tab:single-fin-comparison}}. The DC performance comparison highlights significant improvements in the optimized devices relative to the baseline. At $V_{\text{GS}}$~=~5~V and $V_{\text{DS}}$~=~12~V, device~D2 achieves $I_{\text{DS}\mathrm{max}}$ of 26.91~mA, nearly 10.6$\times$ greater than the baseline (2.52~mA) and substantially higher than device~D1 (17.1~mA). Similarly, the $g_{\text{m}\mathrm{max}}$ is significantly improved with device~D2 reaching 6.84~mS, compared to 1.79~mS and 5.46~mS for, respectively, the baseline and device~D1. Both optimized devices maintain E-mode operation, with $V_{\text{th}}$ ranging from 0.45~V to 0.49~V. It is worth noting that the $W_{\text{eff}}$ in the proposed devices (and particularly in device~D1) is considerably higher than in the baseline, facilitating higher current density. Overall, these results confirm that aggressive structural optimization, when not constrained by expensive simulation costs, can substantially improve $I_{\text{DS}}$ and $g_{\text{m}}$ while preserving thermal stability and threshold voltage integrity.

As expected in wide devices, both device~D1 and device~D2 exhibit increased parasitic capacitance compared to the baseline, including 
gate-to-source capacitance($C_{\text{gs}}$),
gate-to-drain capacitance ($C_{\text{gd}}$),
drain-to-source capacitance ($C_{\text{ds}}$),
total input capacitance ($C_{\text{iss}} = C_{\text{gs}} + C_{\text{gd}}$), and
total output capacitance ($C_{\text{oss}} = C_{\text{gd}} + C_{\text{ds}}$). Notably, device~D1 exhibits a significantly lower input capacitance density than the baseline---an important characteristic for power applications that is expected to yield a lower gate drive loss and thus, higher power efficiency at comparable current levels.

According to {Table~\ref{tab:single-fin-comparison}}, based on the $R_{\text{on}}$ results extracted at $V_{\text{DS}}$~=~0.5~V, it can be seen that device~D2, which has the highest output current, exhibits the lowest value of 37.53~$\Omega$. Both proposed devices demonstrate a significant improvement compared to the baseline device. Additionally, comparing the proposed devices with the baseline device in terms of the power figure of merit (${\text{FoM} = R_{\text{on}} \times (2Q_{\text{G}} + 2Q_{\text{GD}})}$), extracted under small-signal analysis at $V_{\text{GS}}$~=~5~V and $V_{\text{DS}}$~=~6~V, shows that although the proposed circuits have larger $C_{\text{iss}}$ and $C_{\text{oss}}$ capacitances—and consequently more charge—due to their greater fin height ($H_{\text{fin}}$), they still achieve a substantial improvement in FoM as a result of the significant reduction in $R_{on}$. In this case, device~D2 and device~D1, with FoM values of 13.09~p$\Omega\cdot$C and 29.5~p$\Omega\cdot$C, respectively, achieve improvements of approximately 10.8× and 4.79×, respectively, compared to the baseline device, further highlighting the effectiveness of the proposed optimization method.

\begin{table}[htbp]
\caption{\centering Comparison of Key DC and AC Parameters for Single-Fin Baseline and Optimized Devices.}
\label{tab:single-fin-comparison}
\centering
\renewcommand{\arraystretch}{1.2}

    \begin{tabularx}{\columnwidth}{|l|Y|Y|Y|}
        \hline
        \textbf{Parameter}           & \textbf{Baseline\cite{ECTC}} & \textbf{Device~D1} & \textbf{Device~D2} \\ \hline\hline
        $W_{\text{eff}}$ (nm)      & 580                          & 1,555                            & 940.1                            \\ \hline\hline
        \multicolumn{4}{|c|}{\text{DC @ $V_{\text{GS}}$=5~V, $V_{\text{DS}}$=12~V}}                                                   \\ \hline\rowcolor{gray!20}
        $I_{\text{DS}\mathrm{max}}$ (mA/mm) & 4,351                        & 11,013                           & 28,636                           \\ \hline\rowcolor{gray!20}
        $I_{\text{DS}\mathrm{max}}$ (mA)    & 2.52                         & 17.1                             & 26.91                            \\ \hline
        $g_{\text{m}\mathrm{max}}$ (mS/mm)  & 3,102                        & 3,514                            & 7,287                            \\ \hline
        $g_{\text{m}\mathrm{max}}$ (mS)     & 1.79                         & 5.46                             & 6.84                             \\ \hline\rowcolor{gray!20}
        $R_{\text{on}}$ ($\Omega$) & 704.2                        & 86.8                             & 37.53                            \\ \hline\rowcolor{gray!20}
        $V_{\text{th}}$ (V)        & 0.432                        & 0.491                            & 0.448                            \\ \hline\hline
        \multicolumn{4}{|c|}{\text{AC @ $V_{\text{GS}}$=0~V, $V_{\text{DS}}$=6~V}}                                                    \\ \hline\rowcolor{gray!20}
        $C_{\text{gs}}$ (fF)       & 0.3672                       & 0.872                            & 0.790                            \\ \hline\rowcolor{gray!20}
        $C_{\text{gs}}$ (fF/mm)    & 633.1                        & 560.7                            & 840.34                           \\ \hline
        $C_{\text{gd}}$ (fF)       & 0.0501                       & 0.165                            & 0.180                            \\ \hline
        $C_{\text{gd}}$ (fF/mm)    & 86.4                         & 106.1                            & 191.47                           \\ \hline\rowcolor{gray!20}
        $C_{\text{ds}}$ (fF)       & 0.0037                       & 0.001                            & 0.0015                           \\ \hline\rowcolor{gray!20}
        $C_{\text{ds}}$ (fF/mm)    & 6.4                          & 0.643                            & 1.60                             \\ \hline
        $C_{\text{iss}}$ (fF)      & 0.4173                       & 1.037                            & 0.97                             \\ \hline
        $C_{\text{iss}}$ (fF/mm)   & 719.5                        & 667.1                            & 1031.81                          \\ \hline\rowcolor{gray!20}
        $C_{\text{oss}}$ (fF)      & 0.0538                       & 0.166                            & 0.1815                           \\ \hline\rowcolor{gray!20}
        $C_{\text{oss}}$ (fF/mm)   & 92.8                         & 106.7                            & 193.06                           \\ \hline\hline
        \multicolumn{4}{|c|}{\text{AC @ $V_{\text{GS}}$=5~V, $V_{\text{DS}}$=6~V}}                                                    \\ \hline\rowcolor{gray!20}
        $C_{\text{iss}}$ (fF)      & 2.05                         & 4.38                             & 4.05                             \\ \hline\rowcolor{gray!20}
        $C_{\text{oss}}$ (fF)      & 15.04                        & 24.68                            & 25.71                            \\ \hline
        $Q_{\text{G}}$ (fC)        & 10.25                        & 21.9                             & 20.25                            \\ \hline
        $Q_{\text{GD}}$ (fC)       & 90.24                        & 148.08                           & 154.26                           \\ \hline\rowcolor{gray!20}
        FoM (p$\Omega\cdot$C)      & 141.53                       & 29.5                             & 13.09                            \\ \hline
    \end{tabularx}
\end{table}

\begin{table}[tbp!]
    \caption{\centering Comparison Between Multi-Device Configurations Based on the Proposed Devices and EPC Commercial Products.}
    \centering
    \renewcommand{\arraystretch}{1.2}
    \setlength{\tabcolsep}{3.7pt}
    \begin{tabularx}{\linewidth}{|l|cc|cc|c|c|c|}
        \hline
        \multirow{3}{*}{\textbf{Parameter}}  & \multicolumn{4}{c|}{\textbf{ML-Optimised Devices}}              & \multicolumn{3}{c|}{\textbf{EPC Products}}                        \\ \cline{2-8}
        & \multicolumn{2}{c|}{\textbf{Device~D1}$^{\text{\tiny{1}}}$} & \multicolumn{2}{c|}{\textbf{Device~D2}$^{\text{\tiny{1}}}$} & \multirow{2}{*}{\textbf{2040}} & \multirow{2}{*}{\textbf{2216}} & \multirow{2}{*}{\textbf{2023}} \\ \cline{2-5}
        & Switch$^{\text{\tiny{2}}}$ & 50$\times^{\text{\tiny{3}}}$ & Switch$^{\text{\tiny{2}}}$ & 50$\times^{\text{\tiny{3}}}$
        & & & \\ \hline
        $C_\text{iss}$~(pF)
        & $0.18$      
        & $8.86$        
        & $0.16$
        & $8.15$
        & $86$
        & $98$
        & $2{,}150$  \\ \hline
        $C_\text{oss}$~(pF)
        & $0.05$
        & $2.44$
        & $0.04$
        & $1.90$
        & $67$
        & $66$
        & $1{,}530$  \\ \hline
        $Q_\text{G}$~(pC)
        & $5.11$
        & $255$ 
        & $5.24$
        & $262$ & $745$ & $870$ & $19{,}000$ \\ \hline
        $Q_\text{GD}$~(pC) & $0.29$ & $14.6$ & $0.23$ & $11.48$ & $140$ & $130$ & $3{,}200$  \\ \hline
        $Q_\text{G+GD}~\text{(pC)}$ & $5.4$ & $270$ & $5.47$ & $274$ & $885$ & $1{,}000$ & $22{,}200$ \\ \hline
        $R_\text{on}$~(m$\Omega$) & $490$ & $9.8$ & $940$ & $18.8$ & $24$ & $26$ & $1.45$     \\ \hline
        FoM$^{\text{\tiny{4}}}$~(p$\Omega\cdot$C) & $5.29$ & $5.29$ & $10.3$ & $10.3$ & $21.2$  & $26$ & $32.2$ \\ \hline
        $I_{\text{DS}\mathrm{max}}$~(A) & $3.3$ & $165$ & $1.67$ & $83.5$ & $3.4$ & $3.4$     & $90$       \\ \hline
    \end{tabularx}
    \vspace{-2pt}
     \begin{flushleft}
        \footnotesize 
        $^\text{\tiny{1}}$w/o extrinsic parasitics 
        $^\text{\tiny{2}}$300-fin device 
        $^\text{\tiny{3}}$50 parallel switches
        $^\text{\tiny{4}}$lower is better
    \end{flushleft}
    \label{tab:multi-fin-comparison}
\end{table}

\textbf{\textit{Scalable Multi-Fin Configuration.}}
Multi-fin configurations with 300 identical fins designed using device~D1 or device~D2 are designed with filling factors (FF) of, respectively, 46\% and 44\%~\cite{nela2-FF}. A slightly reduced filling factor (compared to 50\% in the baseline) is due to the nature of the single-fin  configuration of these devices compared to the multi-fin structure in~\cite{nela2-FF}. The current $I_{\text{DS}}$ and parasitics $R_{\text{on}}$, $C_{\text{iss}}$, and $C_{\text{oss}}$ are extracted based on TCAD simulations. A fixed fin spacing ($S_{\text{fin}}$) of 100~nm is used for both device designs.

\begin{figure}[t]
    \centering  
    \vspace{-10pt}
    \includegraphics[width=\linewidth]{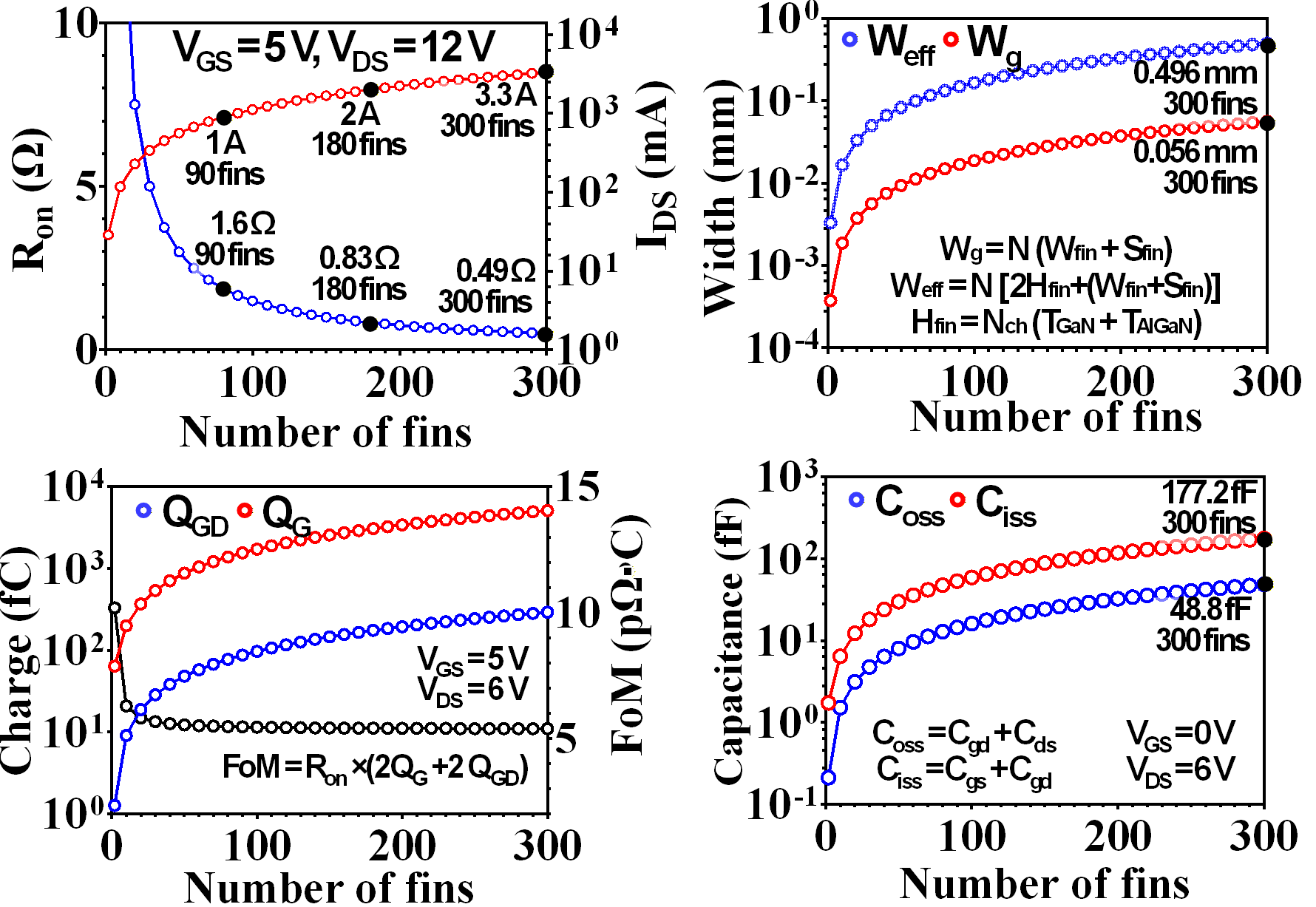}    
    \caption{Design space exploration of device~D1 as a function of the number of fins, showing that increasing fin count reduces $R_{\mathrm{on}}$ while increasing $I_{DS}$, $Q_g$, and capacitances, leading to a trade-off in switching FoM.}
    \label{fig:multiFIN-device1}
\end{figure}

Performance metrics of device~D1 as a function of the number of fins are shown in Fig.~\ref{fig:multiFIN-device1}. As the fin count increases to 300, $I_{\text{DS}}$ rises significantly, exceeding 3.3~A, while $R_{\text{on}}$ drops below 0.5~$\Omega$. This reflects a substantial improvement in current handling and a reduction in conduction losses. Additionally, the $W_{\text{eff}}$ increases more rapidly than the physical gate width ($W_{\text{g}}$), demonstrating the benefits of vertical multi-channel scaling. These trends confirm that fin scaling efficiently improves the power density of tri-gate GaN structures, optimizing both device size and resistive performance in device~D1.

\begin{figure}[t]
    \centering  
    \vspace{-10pt}
    \includegraphics[width=\linewidth]{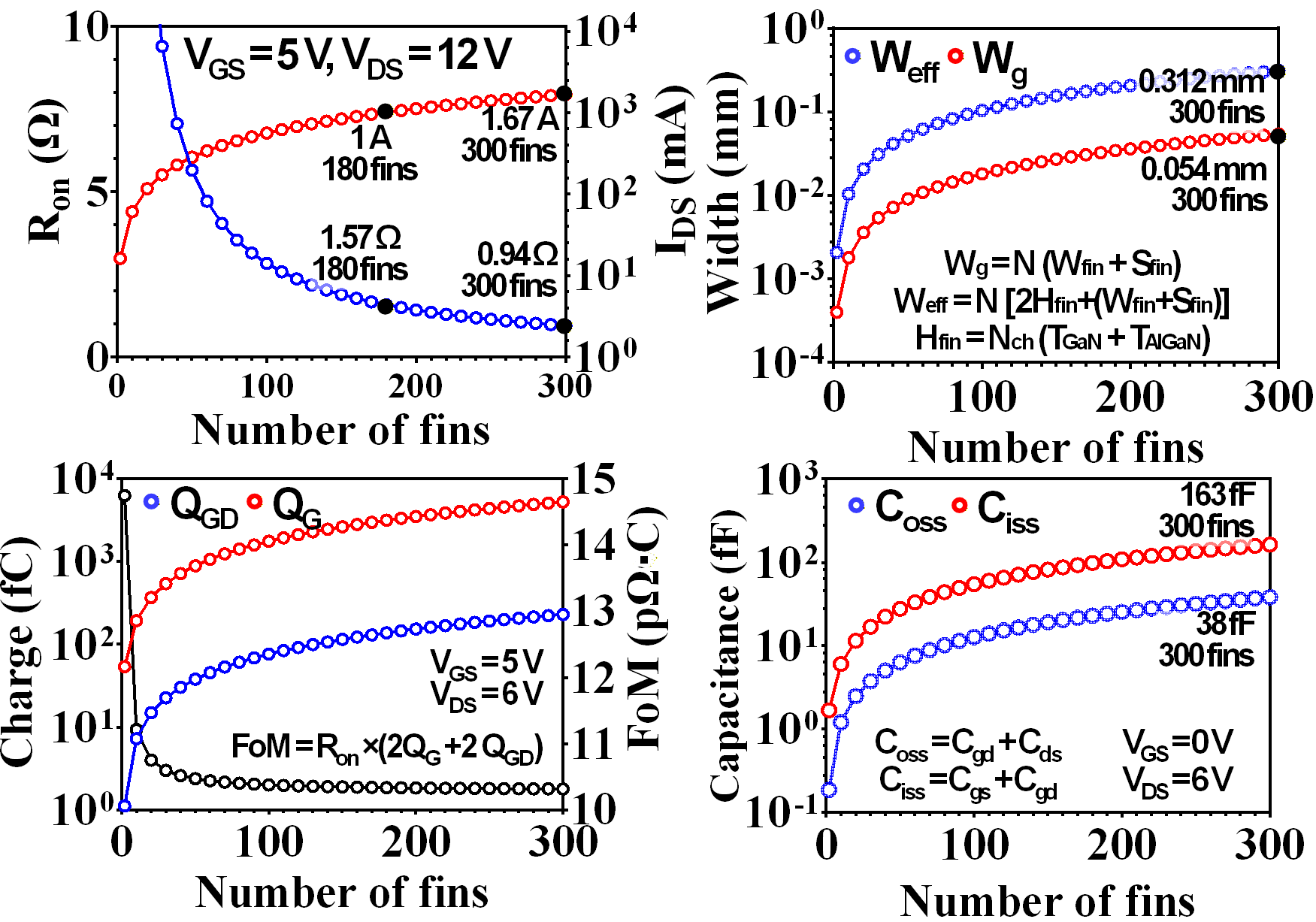}    
    \caption{Design space exploration of device~D2 as a function of the number of fins, showing that increasing fin count reduces $R_{\mathrm{on}}$ while increasing $I_{DS}$, $Q_g$, and capacitances, leading to a trade-off in switching FoM.}
    \label{fig:multiFIN-device2}
\end{figure}

Parasitic capacitance and switching charge behavior for device~D1 are shown in {Fig.~\ref{fig:multiFIN-device1}}. As expected, both $C_{\text{iss}}$ and $C_{\text{oss}}$ increase with the number of fins, but remain within a practical range even at 300 fins. Similarly, the gate charge components $Q_{\text{G}}$ and $Q_{\text{GD}}$ scale with the larger channel area, though their impact is mitigated by the substantial reduction in $R_{\text{on}}$. As a result, the power FoM improves, indicating that device~D1 achieves high current drive with controlled switching losses---a critical balance for high-performance power-switching applications. As shown in {Fig.~\ref{fig:multiFIN-device2}}, device~D2 also benefits from higher number of fins, yet with lower  performance gains compared to device~D1. As the fin count increases to 300, $I_{\text{DS}}$ reaches 1.67~A—nearly half of device~D1's current drive, while $R_{\text{on}}$ is reduced to 0.94~$\Omega$, almost twice as high as the 0.49~$\Omega$ observed in device~D1. This indicates that while device~D2 exhibits superior single-fin performance, it also exhibits current spreading effects when scaled to multi-fin configurations~\cite{main-multifin,nela2-FF,nature-revi}. As a result, its $I_{\text{DS}}$ and $R_{\text{on}}$ are constrained by the thinner GaN channel, thicker AlGaN barrier, and smaller $W_{\text{eff}}$ and $W_{\text{g}}$ compared to device~D1.

In the 300-fin configuration, device~D2 exhibits lower capacitances and switching charges compared to device~D1, with $C_{\text{iss}}$ and $C_{\text{oss}}$ reaching, respectively, 163~fF and 38~fF, as shown in {Fig.~\ref{fig:multiFIN-device2}}. Gate charges $Q_{\text{G}}$ and $Q_{\text{GD}}$ also remain smaller owing to the device's reduced channel volume and gate area. Due to its higher $R_{\text{on}}$, device~D2 achieves a worse FoM despite its favorably lower capacitances. This trade-off between the two devices highlights the critical importance of balancing resistive and capacitive parasitic component, and suggests that device~D1 is better suited for applications that prioritize power efficiency along with maximum current drive.

\textbf{\textit{Scalable Multi-Device Configuration.}}
A system of 50 parallel-connected multi-fin devices is designed and simulated. The results are summarized in {Table~\ref{tab:multi-fin-comparison}} alongside several industrial benchmarks. Both proposed devices demonstrate substantial improvements in switching and conduction performance over the listed commercial benchmarks. Notably, device~D1 achieves $I_{\text{DS}\mathrm{max}}$ of 165~A, nearly double the current capability of the leading commercial device, EPC2023 (90~A), and vastly higher than EPC2040 and EPC2216 (3.4~A each). Furthermore, the device~D1-based 300-fin 50-device system exhibits the lowest power FoM, achieving 5.29~p$\Omega\cdot$C, a nearly 6$\times$ improvement over EPC2023 (32.19~p$\Omega\cdot$C). This remarkable performance results from both its significantly reduced $R_{\text{on}}$ of 9.8~m$\Omega$ and low gate charge.

Alternatively, device~D2 complements these results by exhibiting lower parasitic capacitances, making it more suitable for lower-power voltage regulators where minimizing switching losses is critical. Specifically, device~D2 exhibits ${C_{\text{iss}}~=~8.15}$~pF and ${C_{\text{oss}}~=~1.9}$~pF, both lower than, respectively, the 8.86~pF and 2.44~pF for device~D1 and over two orders of magnitude smaller than those in EPC2023 ($C_{\text{iss}}$~=~2,150~pF, $C_{\text{oss}}$~=~1,530~pF). Despite its smaller capacitances, device~D2  delivers a a high current of 83.5~A, closely matching EPC2023 but at a much lower capacitance and power FoM (10.29~p$\Omega\cdot$C vs. 32.19~p$\Omega\cdot$C).
Importantly, the $Q_{\text{G}}$ and $Q_{\text{GD}}$ of the proposed devices are significantly lower than those of commercial products. The total charge ($Q_{\text{G}} + Q_{\text{GD}}$) of device~D1’s is only 270~pC, compared to 22,200~pC for EPC2023---an almost \text{82$\times$ reduction},enabling faster switching transitions and lower dynamic losses. Overall, the proposed devices, particularly device~D1, achieve high current drive, low on-resistance, and reduced switching losses. These results highlight the potential of optimized multi-fin GaN tri-gate architectures for next-generation power switching.

\begin{figure}[t]
    \centering
    \subfloat{\includegraphics[width=.49\columnwidth]{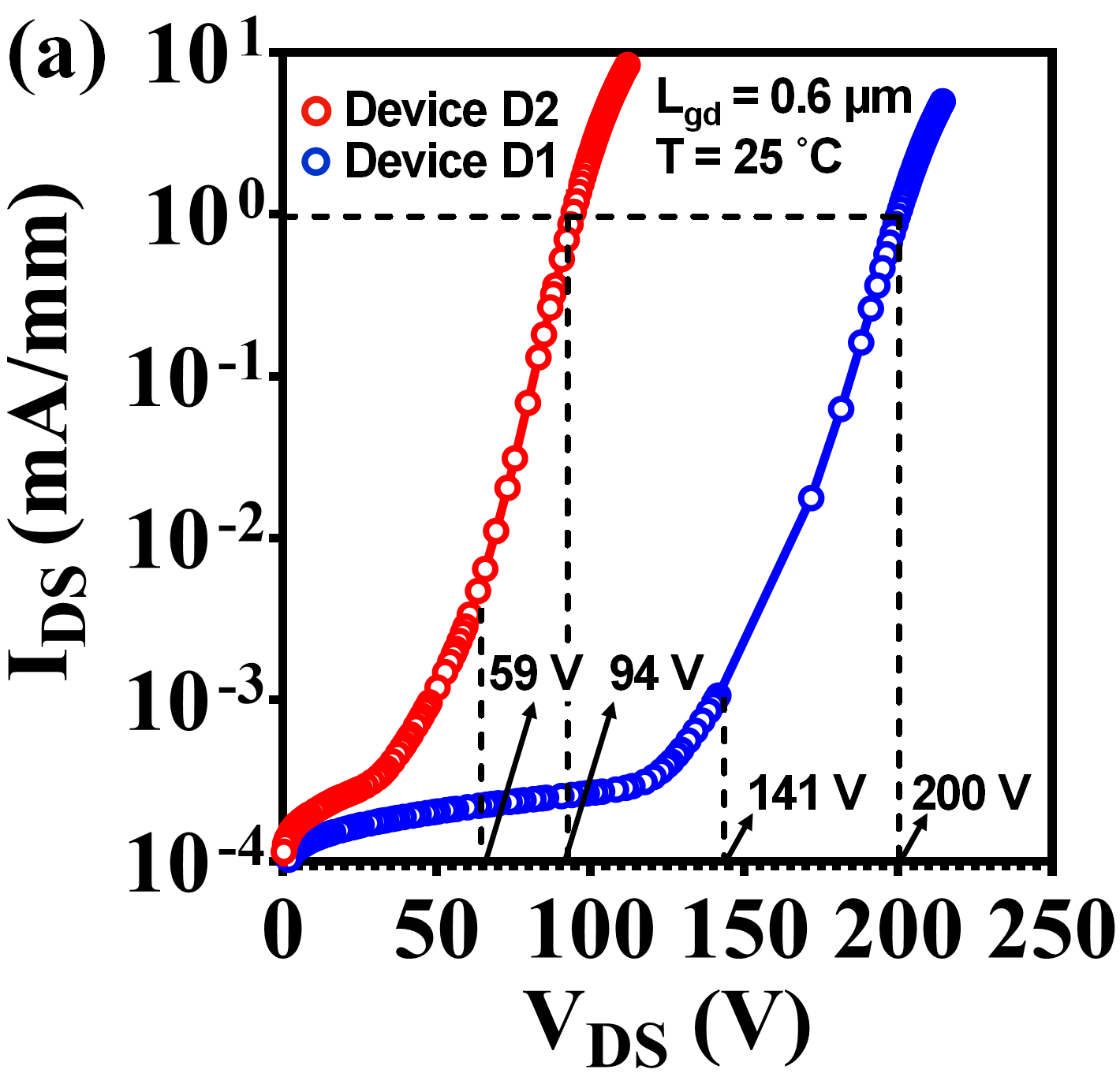}}\label{fig:12a}
    \hfill
    \subfloat{\includegraphics[width=.49\columnwidth]{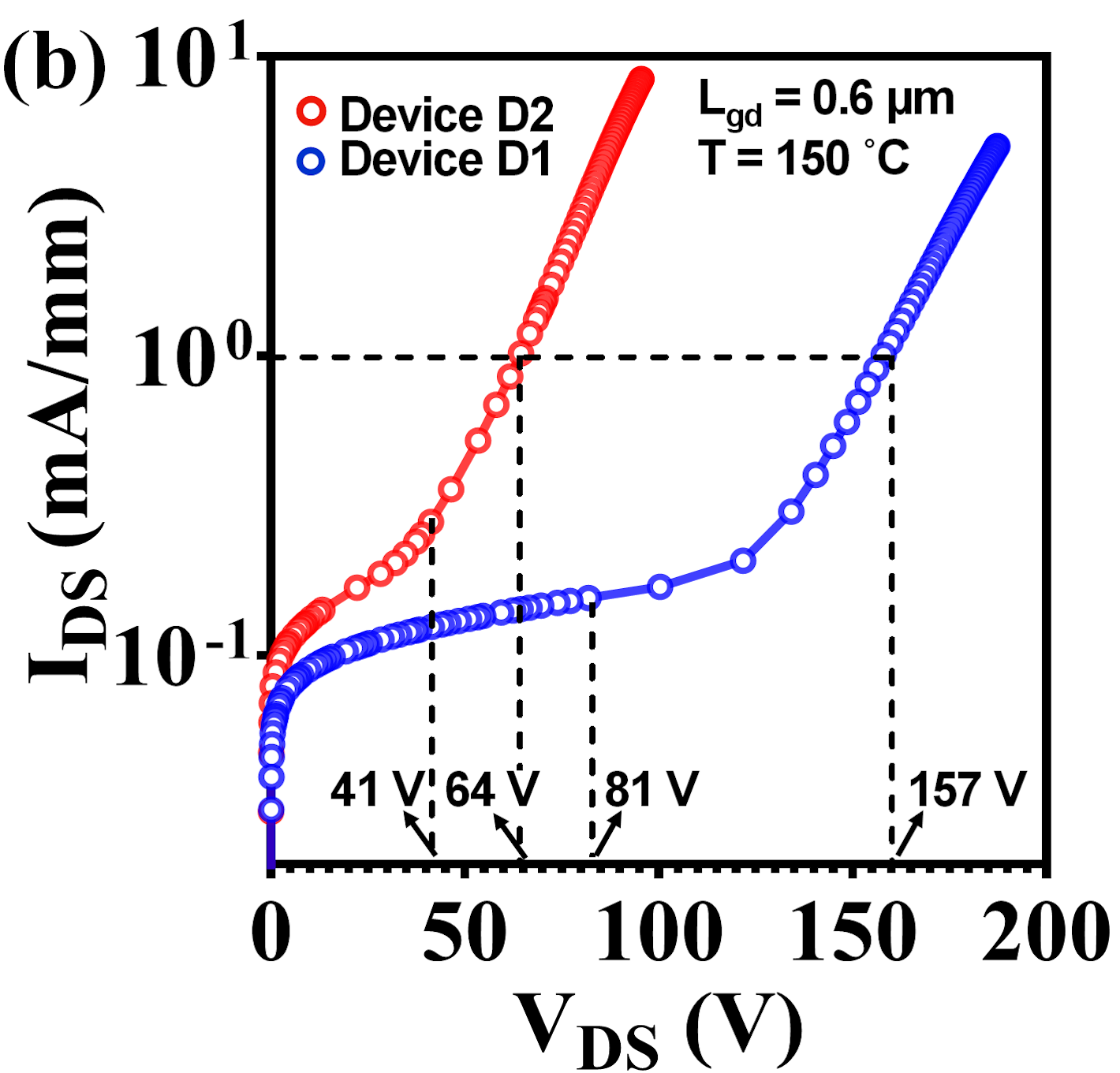}}\label{fig:12b}
\caption{Off-state breakdown characteristics of (a) device~D1 (200 V at 25\textdegree{C}, 157 V at 150\textdegree{C}), and (b) device~D2 (94 V at 25\textdegree{C}, 64 V at 150\textdegree{C}).
}
    \label{fig:Breakdown-Voltage}
    \vspace{-0pt}
\end{figure}

Finally, the off-state breakdown characteristics of device~D1 and device~D2 are shown in {Fig.~\ref{fig:Breakdown-Voltage}(a) and (b)}, extracted at $I_{\text{DS}}$~=~1~mA/mm with $L_{\text{gd}}$~=~0.6~\textmu m under substrate float conditions. At 25\textdegree{C}, device~D1 achieves a breakdown voltage of 200~V, while device~D2 reaches 94~V. At 150\textdegree{C}, these values decrease to 157~V and 64~V, respectively, still maintaining sufficient voltage margins for the targeted applications. These results highlight device~D1’s superior thermal stability and voltage robustness for compact embedded power switches.

\section{Conclusion}
\vspace{-12pt}

Two low-loss, high-current tri-gate FinFET devices with small $R_\text{on}$ are proposed in this paper. The devices are optimized using a physics-informed, active learning-based procedure that effectively reduces computational cost and time of otherwise prohibitive TCAD-based simulations. Compared to prior art and commercial off-the-shelf power switches, the resulting devices demonstrate superior  performance for power delivery applications, paving the way for further optimization of complex device architectures toward application-specific performance targets.

\vspace{-5pt}
\bibliographystyle{IEEEtran}
\bibliography{references}
\vspace{-15pt}

\end{document}